\documentclass[11pt]{article}
\usepackage[letterpaper,margin=1in]{geometry}
\usepackage[T1]{fontenc}
\usepackage{lmodern}
\usepackage{amsmath}
\usepackage{graphicx}
\usepackage{booktabs}
\usepackage{tabularx}
\usepackage{caption}
\usepackage[hidelinks]{hyperref}
\usepackage{url}

\title{Sales Forecasting Using High-Performance Computing}
\author{Sowmya Bhatraju \and Pavan Ghantasala \and Yu Lin Tai \and Ashutosh Porwal \and
Swati Srivastava \and Vineet Suhas Soni \and Yang Wang\\[4pt]
\small Krannert School of Management, Purdue University, West Lafayette, IN 47907, USA}
\date{}

\begin{document}
\maketitle

\begin{center}
\small\emph{Author's version of a paper published in the Proceedings of the 2022 Midwest
Decision Sciences Institute Virtual Conference (Oakland University, April 8--9, 2022),
pp.~622--633.}
\end{center}

\begin{abstract}
This paper aims to improve inventory management and assortment planning of a national
retailer through reliable and efficient sales forecasting. It builds on the ``bottoms-up
approach'' model currently used by the company by creating a robust regression model with
optimal interaction terms, feature engineering, and hyper-parameter tuning using
high-performance computing. Exploratory data analysis was performed on three categories of
data (batteries, brakes, and filters), which provided information about the demographics of
customers, the geography of the store regions, gross revenue, and quantity sold for the past
two years at a unique SKU-store combination. Data aggregation and other transformations were
done as per business requirements and for ease of modeling. Advanced programming was
performed using the Bell cluster and Dask machine learning libraries. Several feature
engineering experiments were carried out to create second-degree interaction terms, PCA
analysis, and so on. The response variable was predicted by building several models such as
linear regression, lasso regression, and OLS. The final regression model selected has the
highest adjusted R-square and least root mean square error, which were the two metrics
defined for model selection. This model also had the highest interpretability and lowest run
time.

\medskip\noindent\textbf{Keywords:} high-performance computing, sales forecasting, feature
engineering, PCA, machine learning
\end{abstract}

\section{Introduction}

``The future belongs to those who prepare for it today'' \cite{malcolmx}. This phrase holds
for all companies that aspire to generate promising sales outcomes in uncertain times. Sales
forecasting is a vital process for achieving those outcomes, as it predicts an
organization's growth and overall health. According to a Forbes article, ``It's a tough job,
but without a sales forecast, the management team doesn't know whether to conserve cash
aggressively or continue current operations. It's not an easy problem to solve, but some
insights are possible'' \cite{conerly}. An organization's ability to forecast sales
accurately is crucial for sustaining itself in the market, competing with its strategic
vision, and operating effectively. In addition, it helps reduce inventory management costs
through better sales predictions, better inventory allocation across stores, and proper
assortment planning.

Accurate sales forecasting is still a significant challenge for most companies. As reported
in the Wall Street Journal, ``It's been an adventure over the past year,'' said Hayley Berg,
head of price intelligence at travel-booking app Hopper Inc. ``It's really a challenging
forecasting problem,'' she said, referring to the abrupt halt in consumer and business travel
the previous year \cite{loten}. Without a precise sales forecast, organizations can harm
their estimates over a long-term horizon. In a Gartner press release, ``Heads of sales
operations are under constant pressure to produce accurate forecasts to help shape decision
making,'' said Craig Riley, senior principal analyst in Gartner's Sales Practice.
``Unfortunately, sales forecasting isn't getting easier --- as expanding product portfolios
and shifting market conditions exacerbate the issue'' \cite{blum}. The 2019 Gartner State of
Sales Operations Survey cited in that release (sample size 296) found that fewer than half of
sales leaders and sellers have high confidence in their organization's forecast accuracy
\cite{blum}. All in all, sales forecasting has the potential to make or break a business.

This paper revolves around two primary research questions.

\paragraph{How can sales be forecast using different linear regression techniques, including
interaction terms and feature engineering?}
Sales can be forecast by identifying predictor input variables for the machine learning
model, then manipulating and transforming those input variables into features that can boost
the model's prediction accuracy, performance, and efficiency. At times there is an
interaction among the predictor variables, indicating a relation between input variables that
affects the response. Therefore, poor feature engineering, or failing to acknowledge
interactions among variables, can harm a sales forecasting model.

\paragraph{How would using HPC capabilities allow us to find an optimal robust linear model
that performs much better than a generic backward-selection approach?}
High-performance computing (HPC) systems can solve complex computational problems swiftly
through enhanced processing capability and storage capacity. HPC can help create robust
models that take advantage of vast volumes of data with higher computing and performance
specifications, which cannot be done through mere feature selection. It uses dense computer
clusters that synchronize with one another, allowing a model to run on optimized hardware
with advanced processing capabilities. With these capabilities, a model with optimal
interaction terms, feature engineering, and hyper-parameter tuning using high-performance
computing can perform sales forecasting with higher interpretability and lower run time.

The remainder of this paper is organized as follows. Section~\ref{sec:related} reviews the
literature on the methods used to date and explains how our analysis addresses a gap in it.
Section~\ref{sec:data} describes the store-level SKU data from the retailer.
Section~\ref{sec:method} presents the proposed methodology, Section~\ref{sec:model} the
models, and Section~\ref{sec:results} their performance. Section~\ref{sec:conclusion}
concludes with the implications of this study and directions for future research.

\section{Related Work}\label{sec:related}

\subsection*{Dealing with high dimensionality and computation issues}

Sales forecasting is the process of predicting the sales a firm will generate within a
specific timeframe such as a week, month, quarter, or year. Creating a predictive model to
forecast sales with high dimensionality requires careful consideration so that model
computation time stays within an acceptable time frame. Sales forecast modeling with high
dimensionality in the retail domain can be approached in different ways. When there are many
explanatory variables, multiple variable selection methods can be combined, and the union of
the variables they select can be taken to reduce the risk of eliminating variables with a
significant impact on the response variable (Stock and Watson, 2004). PCA can be performed to deal
with high dimensionality, but it runs the risk of eliminating features that might be strong
predictors of the dependent variable. One way around this is to group variables by their
common attributes and run PCA on each group separately, reducing the likelihood of missing
important predictors \cite{ma2016}. Evolutionary algorithms are a less-known approach to
feature selection in high-dimensional data. They are inspired by the principles of evolution:
new candidate solutions are created by transforming the best-performing past candidates.
Although practical, evolutionary algorithms demand high computing power on high-dimensional
data, so high-performance computing has been employed to apply them to feature selection in
high-dimensional datasets \cite{alwadei}.

There is substantial literature on causal forecasting that assumes a functional form for the
effect of different explanatory variables on sales. Fader and Hardie \cite{fader} argued that
consumer choices and demand for SKUs are not formed at an individual level; instead, a set of
discrete attributes such as package, size, and type can characterize future demand. The
marketing literature has augmented sales prediction models with promotional activities as
predictors, for example discounts \cite{cooper}, the frequency of promotions for similar
products \cite{christen}, and category or product group \cite{narasimhan}. Trapero,
Kourentzes, and Fildes \cite{trapero} proposed promotional modeling that combines PCA and
regression analysis at the SKU level. Retail demand estimation has historically been
dominated by consumer choice models, focusing on likelihood estimation and
expectation-maximization algorithms with multinomial logit models and utility functions
\cite{vulcano,kok,anupindi}. Most of these models are based on stock-out event data
\cite{ma2016}. However, there is evidence that demand depends on the inventory level for at
least some products, with higher inventory levels leading to higher sales, the so-called
``billboard effect'' \cite{koschat,ton}.

Machine learning models such as decision trees, support vector machines, or neural networks
do not assume such direct functional relationships, but their complexity makes them
computationally expensive in a big data environment. Linear regression has been the optimal
choice for practitioners balancing interpretability, time complexity, and accuracy in sales
prediction. It is easy, simple, and feasible to fit in a high-dimensional SKU-level space,
and several approaches can capture non-linearities, such as multiplicative interactions and
log and exponential transformations of the variable space \cite{ma2016}. Divakar et
al.~\cite{divakar} designed a dynamic regression model to capture the effects of past sales,
trends, and regional factors.

\section{Data Description}\label{sec:data}

The data used in the project was provided by a national auto-parts dealer and is divided into
three categories: filters, brakes, and batteries. Fields are consistent across the three
categories, with over 80 predictors. Each record refers to a unique SKU-store combination.
More than 15 features describing the demographics and geography of the store regions are
included (population density, road quality index, average age, percentage of blue-collar
workers, median household income, vehicles in operation in the region, and so on). Features
on sales in the past two years (gross revenue, net revenue, quantity sold, and so on) are also
included. Stores have been clustered into groups based on sales performance and location, and
average cluster performance (average cluster unit sales, average cluster lost sales, and so
on) is also used as input. Other predictors include missed sales, lost quantity of an item,
the number of times an SKU is looked up in the catalogue, and the life cycle of the SKU.
Separate tables for inventory, sales, SKUs, and stores were also received; their data
dictionaries are given in Tables~\ref{tab:inventory}--\ref{tab:store}.

\begin{table}[htbp]
\centering\small
\caption{Inventory}\label{tab:inventory}
\begin{tabular}{lll}
\toprule
Variable & Type & Description\\
\midrule
storenum & Numeric & Store number\\
skunum & Numeric & SKU number\\
inv\_year & Numeric & Year of inventory\\
inv\_period & Numeric & Period of inventory\\
wk\_of\_period & Numeric & Week of period\\
wk\_of\_year & Numeric & Week of year\\
smaxi & Numeric & Maximum; if smaxi $>0$ the SKU is stocked at the store\\
scoh & Numeric & Quantity of units on hand in store\\
wk\_ending\_dt & Date & Week ending date\\
merchandise\_group\_desc & Text & Merchandise group description (batteries, etc.)\\
\bottomrule
\end{tabular}
\end{table}

\begin{table}[htbp]
\centering\small
\caption{Sales}\label{tab:sales}
\begin{tabular}{lll}
\toprule
Variable & Type & Description\\
\midrule
store\_number & Numeric & Store number\\
sku\_number & Numeric & SKU number\\
customer\_type & Text & Type of customer (DIFM)\\
qty\_sold & Numeric & Quantity of SKUs sold\\
gross\_sales & Numeric & Gross sales for the fiscal year\\
sales\_cost & Numeric & Cost to company for the fiscal year\\
fiscal\_year & Date & Fiscal year\\
fiscal\_period & Numeric & Fiscal date\\
\bottomrule
\end{tabular}
\end{table}

\begin{table}[htbp]
\centering\small
\caption{SKU}\label{tab:sku}
\begin{tabular}{lll}
\toprule
Variable & Type & Description\\
\midrule
sku\_number & Numeric & SKU number\\
mpog\_description & Text & Master Plan of Gram description\\
sku\_description & Text & Description of SKU\\
length & Numeric & Length of SKU\\
width & Numeric & Width of SKU\\
height & Numeric & Height of SKU\\
cubic\_inches & Numeric & Volume of SKU\\
retail\_price & Numeric & Retail price of SKU\\
unit\_cost & Numeric & Cost per unit\\
min\_app & Numeric & No.\ of applications needed to complete the project\\
discontinued\_flg & Categorical & SKU discontinued or not\\
stocking\_location & Text & Location of the stock\\
epc\_sku\_type\_description & Text & Type of SKU\\
store\_acquisition\_cost & Numeric & Store acquisition cost\\
\bottomrule
\end{tabular}
\end{table}

\begin{table}[htbp]
\centering\small
\caption{Store}\label{tab:store}
\begin{tabular}{lll}
\toprule
Variable & Type & Description\\
\midrule
store\_number & Numeric & Store number\\
dma\_id & Numeric & Designated Market Area ID\\
sector\_id & Numeric & Sector ID\\
market\_class & Categorical & Market class\\
latitude & Numeric & Latitude of store\\
longitude & Numeric & Longitude of store\\
platform\_cluster\_name & Categorical & Platform cluster name\\
platform\_cluster\_name\_py & Categorical & Platform cluster name previous year\\
open\_date & Date & Store opening date\\
market\_share & Numeric & Market share\\
capture\_rate & Numeric & Ratio of sales in per store, sector and sales per store\\
rolling\_capture\_rate & Numeric & Rolling capture rate\\
\bottomrule
\end{tabular}
\end{table}

\section{Methodology}\label{sec:method}

The sales forecasting approach is formulated as a regression problem, with the optimal
prediction judged on root mean square error (RMSE). Because such an optimal solution depends
heavily on computational capacity, the linear regression model's potential is maximized by
using parallel computing and the enhanced storage and performance of high-performance
computing (increasing nodes). The overall methodology has four parts: exploratory data
analysis, data pre-processing, feature extraction and feature engineering, and model building
with performance benchmarking.

The sales data lists SKU- and store-level information documented as monthly aggregates and
yearly trends. The SKUs under consideration belong to three categories (brakes, filters, and
batteries), which make up the bulk of the auto-parts retailer's sales. The exploratory data
analysis aimed to summarize the data as a whole, extract as many insights as possible, and
discover patterns and anomalies. It was also a step for investigating potential interactions
and their significance in the regression problem.

Of the 12,121 unique SKUs identified, 40\% were heavy-duty application-specific. In terms of
profitability, brake pads brought in the most revenue, whereas brake shoes had lower profit
margins. There were 1,207 SKUs discontinued across stores, and negative sales implied a need
for imputation. This step therefore set the foundation for specific pre-processing, such as
scaling, imputation, and outlier handling, to improve prediction capability.

The data pre-processing stage reduced unwanted information through data reduction,
transformation, and identification of potential interactions between variables. Data
reduction eliminated low-business-impact variables and null and missing occurrences: all
current-year sales data, part types, cluster, adjusted vehicles in operation, and
failure-specific information were removed. Since actual current-year sales introduce bias,
removing these columns helped ensure greater prediction accuracy and reduced non-predictors
in the model. All SKUs with zero sales were considered insignificant for accurate prediction
and removed from the dataset. Several variables, such as actual vehicles in operation,
establishments, and road quality indices, were imputed with central aggregates, and negative
sales were replaced with zero to avoid skew. Further transformations included creating
dummies for categorical variables so that multiple groups could be used in the regression,
scaling and standardizing the data, and checking the correlation matrix to identify useful
variable interactions.

To increase the linear regression's efficacy and use the insights from the correlation map,
we created selected interactions. These were second-order terms, which bring in first and
second powers of the parameters (equation~\ref{eq:inter}). The aim was to find variables that
had a positive business impact on the overall score and MSE of the regression model,
increasing prediction accuracy. Equations~(\ref{eq:nointer}) and~(\ref{eq:inter}) show the
difference between the models without and with interaction terms. Interaction terms capture
synergies and the combined impact of two variables on sales. Interaction terms were chosen on
the basis of observed statistical significance. It was also important to analyze terms that
were insignificant on their own but raised the statistical significance of the overall
prediction as part of an interaction, which makes it possible to identify variables with an
intrinsic impact on sales. Interaction variables can also demonstrate business impact: for
example, a history of good sales for an SKU indirectly implies that the item is available in
that store.
\begin{align}
\text{Sales}_{\text{no interactions}} &= \beta_1 x_1 + \beta_2 x_2 \label{eq:nointer}\\
\text{Sales}_{\text{with interactions}} &= \beta_1 x_1 + \beta_2 x_2 + \beta_{11} x_1^2
 + \beta_{22} x_2^2 + \beta_{12} x_1 x_2 \label{eq:inter}
\end{align}

\begin{figure}[htbp]
\centering
\includegraphics[width=\textwidth]{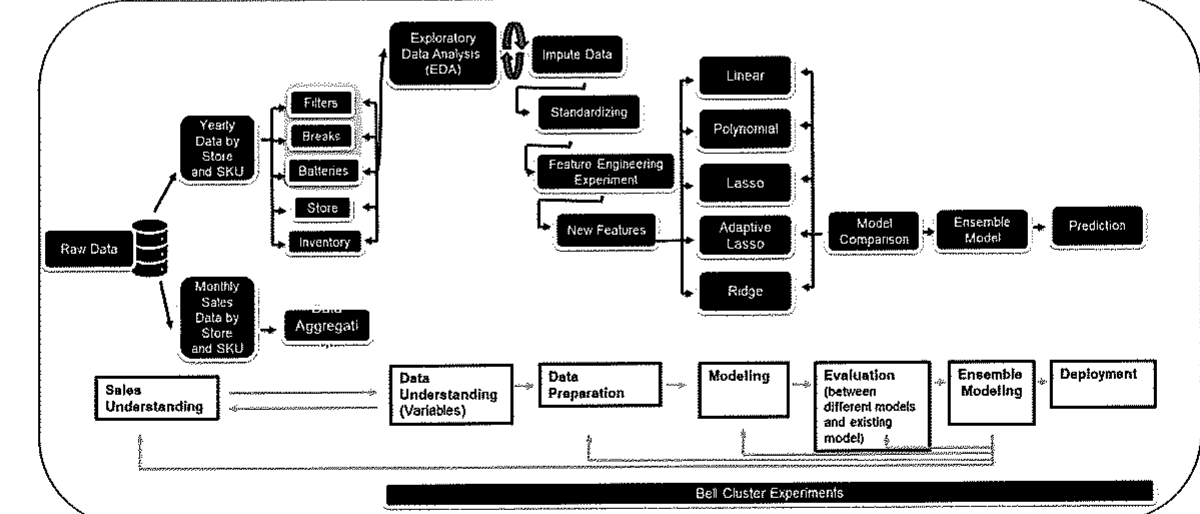}
\caption{Study design.}\label{fig:design}
\end{figure}

Figure~\ref{fig:design} summarizes the study design.

\section{Model}\label{sec:model}

\subsection{Libraries and functions}
The Dask-ML library was used for modeling. It provides scalable, distributed machine learning
for the challenges of large data size, high dimensionality, and computation. Dask breaks the
dataset into small chunks, which parallelizes tasks and reduces computational load. Every
function run on Dask dataframes is stored in a workflow and not computed until explicitly
called. This ``lazy'' computation greatly reduces the computational requirement for operations
on large datasets. The library provides its own functions such as \texttt{make\_regression},
\texttt{StandardScaler}, \texttt{PolynomialFeatures}, and \texttt{PCA}. Dask dataframes were
used for data pre-processing, and the processed Dask dataframe was converted back into a
pandas dataframe to run the model.

\subsection{Data transformation}
Several transformations were carried out on the data as per business requirements and for
ease of modeling. Stores with outlier sales were dropped. To perform feature engineering with
interaction terms, variables with null values were imputed using business sense, and some
null values were dropped. Current-year variables were not used in the model, as they are not
predictors. Null values for part type were replaced with ``Miscellaneous''. SKUs with zero
previous-year gross sales were dropped, since gross sales is a significant predictor. The data
was normalized using standard scaling.

\subsection{Feature engineering}
Second-degree interaction terms were created, and principal component analysis (PCA) was
carried out. PCA projected the data into a lower-dimensional space, reducing linear
dimensionality using singular value decomposition. The number of components was varied
depending on the variance explained. The quantity of SKUs sold in the current year was the
response variable.

\subsection{Data modeling}
\begin{enumerate}
\item \textbf{Linear regression.} Linear regression is simple to implement, and its output
coefficients are easy to interpret. It is less complex than other algorithms and hence easier
to use. It can over-fit, but this can be avoided through dimension reduction, regularization,
and cross-validation.
\item \textbf{Lasso regression.} The Least Absolute Shrinkage and Selection Operator (LASSO)
is a type of linear regression that uses shrinkage in the loss function and automates feature
selection. It is useful when the predictors have a mixture of large and small coefficients.
\item \textbf{OLS.} The OLS estimator is reliable when the Gauss--Markov (OLS) assumptions are
met. If the data fulfil all the required assumptions, OLS is the best linear unbiased
estimator (BLUE).
\end{enumerate}

\section{Results}\label{sec:results}

\begin{figure}[htbp]
\centering
\includegraphics[width=0.8\textwidth]{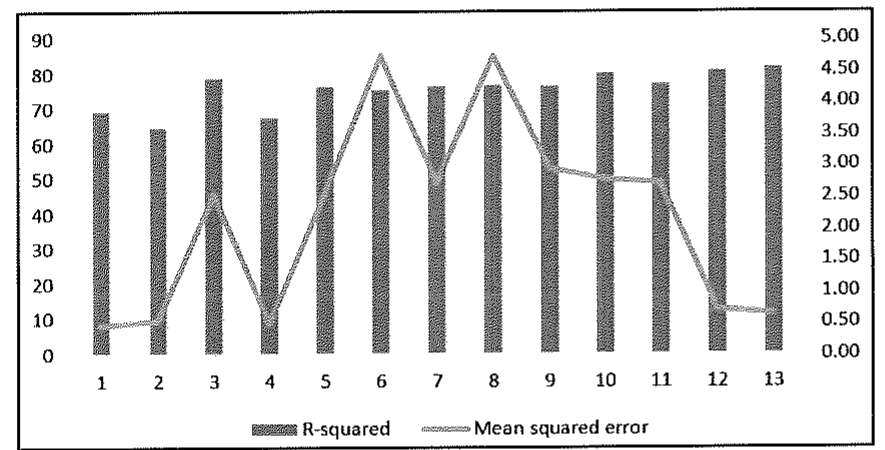}
\caption{Model results: R-squared (bars, left axis, \%) and mean squared error (line, right
axis) for models 1--13.}\label{fig:results}
\end{figure}

Figure~\ref{fig:results} shows the best experimental models, with high R-square and low mean
squared error. R-square ranged from 64.6\% to 82\%, and MSE from 0.465 to 4.722. Linear
regression (models 1, 2, 12), lasso (model 4), and OLS (models 3, 5--11, 13) were explored.
Interaction terms were created by combining demographic variables, lost sales and quantities
in the previous year, and quantity sold in the previous and the year before the previous
year. PCA was also performed. Model 1 (linear regression) had the minimum MSE of 0.71, but its predictors
were not significant, which led to a lower R-square. A few OLS models (6, 8) had a decent
R-square of about 75\% but a high MSE of 4.72, indicating a highly biased or high-variance
estimate. The best model was a linear regression model (12) that used the interaction between
quantity sold in the previous year and the year before that, with an R-square of 81\% and an
MSE of 0.71. The mean absolute error averaged about 0.186.

Similar results were observed on the brakes data: linear regression, lasso regression, and OLS
gave R-square values from 0.81 to 0.835, with mean squared errors of 0.62 to 0.707. On the
batteries data, an R-square of 0.911 was observed with interaction terms, implying a better
model fit.

\section{Conclusion}\label{sec:conclusion}

Using high-performance computing on the Bell cluster, we were able to run hundreds of
experiments with different models and reached an MSE below 1 with a fairly good R-square above 90\%. Our model could create a strong business impact for the company, with the
following monetary benefits:
\begin{itemize}
\item By decreasing mean absolute error by about 90\%, sales could be predicted more
accurately, which would further reduce the company's inventory management costs.
\item With model run time as low as 2 minutes, it offers another cost-saving opportunity.
\end{itemize}

\subsection*{Future scope}
Since parallel computing at this scale is computationally demanding, more interactions can be
incorporated through parallel computing to enhance the predictive power of the regression
model.

\end{document}